\documentclass{article}

\usepackage{arxiv}

\usepackage[utf8]{inputenc} 
\usepackage[T1]{fontenc}    
\usepackage{hyperref}       
\usepackage{url}            
\usepackage{booktabs}       
\usepackage{amsfonts}       
\usepackage{nicefrac}       
\usepackage{microtype}      
\usepackage{lipsum}
\usepackage{graphicx}
\usepackage{lettrine}
\usepackage{caption}
\graphicspath{ {./images/} }
\usepackage{cite}
\usepackage{verbatim}
\usepackage{amsmath}
\usepackage{amssymb}
\usepackage{caption}
\usepackage{geometry}
\usepackage{tikz}
\usepackage{bm}
\usetikzlibrary{shapes,arrows}
\usepackage{proof}
\usepackage{pgf}

\title{Holonorm}

\author{
 Daryl Noupa Yongueng,  Hamidou Tembine\\
  School of Engineering\\
  Electrical Engineering and Computer Engineering Department \\
  Quebec University at Trois-Rivières\\
  \texttt{darly.noupa.yongueng@uqtr.ca, tembine@ieee.org} \\
}

\begin{document}
\maketitle
\begin{abstract}
Normalization is a key point in transformer training . In Dynamic Tanh (DyT), the author demonstrated that Tanh can be used as an alternative layer normalization (LN) and confirmed the effectiveness of the idea. But Tanh itself faces orthogonality, linearity and distortion problems. Due to that, his proposition cannot be reliable. So we propose a Holonorm (hn) which has residual connections and nonlinearity. Holonorm is suitable for replacing Tanh in the context of normalization. Although the HoloNorm expression could be similar to the softsign function in dimension one, softsign is a componentwise function which is not good for tensors and vectors of great dimension. Holonorm preserves the orthogonality, the direction, the invertibility of the signal. Holonorm is also a suitable metric, maps all vectors into the open unit ball. This prevents exploding activations and improves stability in deep Transformer models. In this work, we have meticulously examined the normalization in transformers and say that Holonorm, a generalized form of softsign function suited as a normalization function first.Second, defined between 0 and 1 hn serves as a percentage, and $1 - \text{Holonorm}$ is its complement, making it better understandable in evaluating a model. 
\end{abstract}


\section{Introduction}
	\lettrine{F}inding the best method to train deep learning networks is a long-standing challenge. Layer normalization has became a fundamental aspect in machine learning model architecture. Scientists have proposed several expression to enhance the perfomance of a model for a specific task. In the literature, we have the Gaussian error linear unit (Gelu) \cite{lee2023gelu} that surpasses Rectify error linear unit (Relu) \cite{mastromichalakis2020alrelu}. But one of the most popular is Tanh \cite{zhu2025transformers} which got some symetric properties that make it very intersting but not enough in our point of view because of his limitations. \\
	
	In Hilbert spaces, applying componentwise means possibly an infinite number of coefficients. \texttt{tanh} is computationally ineffective  compared to the calculation of a norm. A striking point of Tanh is that it strongly correlates certain vectors that are uncorrelated at the beginning of the training. When it comes to orthogonality between words (uncorrelated words), by applying \texttt{tanh}, it can make them correlated. But by applying \textbf{holonorm}, it stays uncorrelated. Holonorm keeps them uncorrelated if they were uncorrelated initially. Thus, Tanh does not maintain orthogonality and therefore is not suitable for tensors, vectors then textual data \label{1}.\\
	
	\subsection*{Literature review}
	
	Layer normalization is a technique used in neural networks to normalize the inputs across the features for each data point. Normalization plays a fundamental role in stabilizing training and improving the convergence of deep neural networks \cite{ba2016layer}. While various normalization techniques have been proposed, the use of Tanh-based normalization introduces several challenges in training Transformer architectures. The Tanh activation function, although bounded and smooth, suffers from saturation effects where the gradients become vanishingly small for large input values \cite{glorot2010understanding}. This leads to decrease in gradient flow in deep networks, significantly slowing down convergence and altering learning in early layers~\cite{bengio1994learning}.
	
	In Transformer models, which rely heavily on residual connections and multi-head attention mechanisms, such vanishing gradients can accumulate across layers, causing optimization instability~\cite{vaswani2017attention}. Moreover, Tanh outputs are symmetric around zero, but the compressive nature of the function near its limits leads to information loss in hidden states~\cite{he2015delving}. These issues are particularly pronounced in sequence-to-sequence tasks where long-term dependencies are essential, making Tanh less suitable compared to other normalization techniques such as LayerNorm or recent adaptive methods \cite{xiong2020layer}.
	Recent studies emphasize the importance of non-saturating activation and normalization mechanisms to preserve representational fidelity and training efficiency. As such, replacing Tanh in normalization schemes has become critical in addressing depth-related degradation in Transformers.
	
\subsection*{Holonorm}	
	Holomorphic normalization (HoloNorm) is an activation function better than Hyperbolic Tangent (Tanh) to normalize a function during the training of a machine learning model due the relevant facts mention earlier \ref{1}. 
	Holonorm is defined as: 
\[
\text{HN}_p(x) = \frac{x}{\left(1 + \|x\|_p\right)}
\]
	Where p is the range of the norm, typically \( p = 1 \) for norm 1 and \( p = 2 \) for euclidiean norm. The function is defined for all \( x \in \mathbb{R}^D \) and maps to 
	the open unit ball in \( \mathbb{R}^D \).
	
	In norm-1, softsign function is assimilated to holonorm but holonorm is defined in the open unit ball. 
	\begin{table}[h!]
		\centering
		\begin{tabular}{p{6cm}p{6cm}}
			\includegraphics[width=0.2\textwidth]{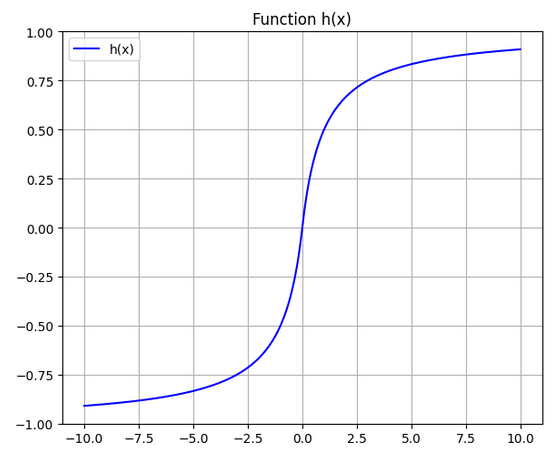} & \includegraphics[width=0.2\textwidth]{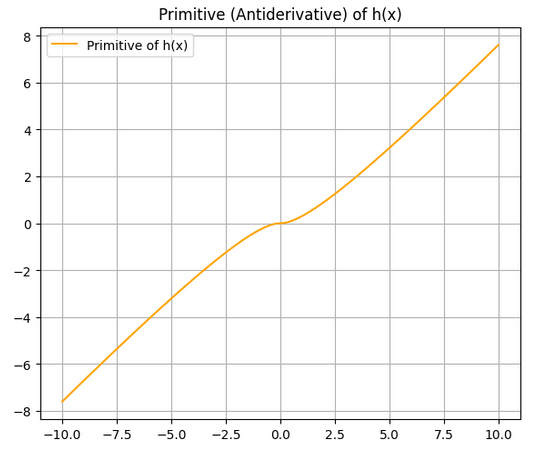} \\
			\captionof{figure}{Curve of $hn(x)=\frac{x}{1+\|x\|}$} & \captionof{figure}{Primitive  $H(x)$:} \\
			&
			\[
			H(x) = \begin{cases}
				x - \ln(1 + x) & x \geq 0 \\
				x + \ln(1 - x) & x < 0
			\end{cases}
			\] \\
			\includegraphics[width=0.2\textwidth]{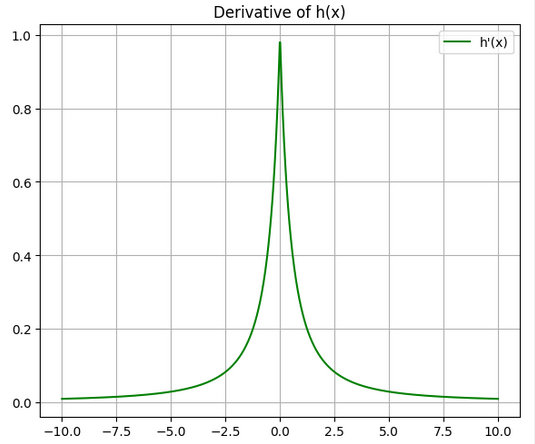} & \includegraphics[width=0.2\textwidth]{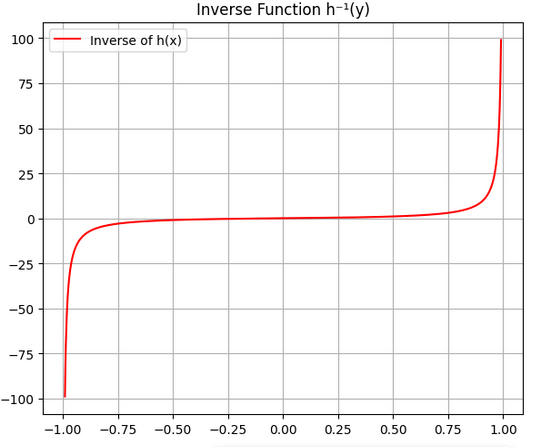} \\
			\captionof{figure}{Derivative of $h'(x) = \frac{1}{(1 + |x|)^2}$} & \captionof{figure}{Inverse of $h^{-1}(y) = \frac{y}{1 - |y|}, \quad y \in (-1, 1)$}
		\end{tabular}
		\caption{Holonorm and her correlated curves}
		\label{tab:images}
	\end{table}
	
	This transformation is scale-sensitive but direction-preserving, unlike tanh, it scales the entire vector uniformly rather than compressing individual components. 
	It preserves the direction by approximately maintaining the angle between vectors and keeps orthogonal signals nearly orthogonal after transformation. Additionally, it does not distort unrelated signals, as independent vectors remain uncorrelated, making it suited for reconstruction since the original signal geometry is largely preserved, allowing for better recovery. 
	
	Considering the metric evaluation, it does the ratio between the actual value and the predicted value or vice versa. Either if the actual value is zero,  the denominator will never be zero, then output will not tend toward infinity, allowing the reconstruction of the initial terms.

	\subsection*{Contribution}
	In this paper, our aim is to propose a novel function that can serves for differents operations as a normalization, activation, and evaluation function. 
	In section 2 we will present advantages of Holonorm on Tanh and the reasons why we should replace it by Holonorm. Section 3 we will do and experiment to prove our statement with a discussion and the last section is the conclusion. 
	
	\section{Reason why HoloNorm is better than Tanh}
	The mathematical properties of Tanh and HoloNorm are compared in Table \ref{tab:comparison}. The general Holonorm hn(x) function is defined as:
	
	\textbf{Proposition 1:} Holonorm is mathematical better than Tanh in every use case \label{2}.

	\begin{tabular}{|p{3cm}|p{3.5cm}|p{3.5cm}|p{3.5cm}|}
		\hline
		\textbf{Property} & \textbf{Tanh} & \textbf{Holonorm} & \textbf{Remarks} \\
		\hline
		Expression & $\tanh(x) = \dfrac{e^x - e^{-x}}{e^x + e^{-x}}$ & $\text{HN}(x) = \dfrac{x}{1 + |x|}$ & Holonorm is simpler \\
		\hline
		Inverse & $\tanh^{-1}(y) = \dfrac{1}{2} \ln\left(\dfrac{1 + y}{1 - y}\right)$ & $\text{HN}^{-1}(y) = \dfrac{y}{1 - |y|}$ or $\dfrac{y}{1 + |y|}$ & Holonorm is faster, easier \\
		\hline
		Derivative & $\dfrac{d}{dx} \tanh(x) = 1 - \tanh^2(x)$ & $\dfrac{d}{dx} \text{HN}(x) = \dfrac{1}{(1 + |x|)^2}$ & Holonorm is cheaper \\
		\hline
		Integral & $\int \tanh(x) \, dx = \ln(\cosh(x)) + C$ & $\int \text{HN}(x) \, dx = \begin{cases} (1 + x) - \ln(1 + x) + C & \text{if } x \geq 0 \\ x + \ln(1 - x) + C & \text{if } x < 0 \end{cases}$ & Holonorm integrable in parts \\
		\hline
		Numerical Stability & Poor near $x \gg 1$ (saturates) & Stable everywhere & Holonorm is more stable \\
		\hline
		Computational Cost & Involves $\exp(x)$ and $\ln(x)$ & Simple algebraic ops only & Holonorm is more efficient \\
		\hline
	\end{tabular}
	
	\begin{figure}[h]
		\centering
		\includegraphics[width=0.5\textwidth]{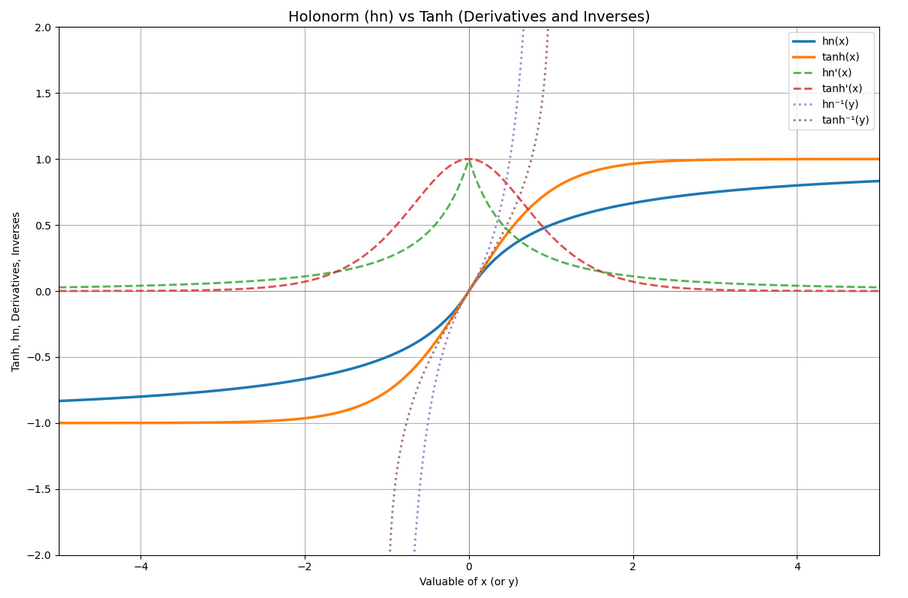}
		\caption{Differents curves of holonorm and Tanh}
		\label{fig:hn_vs_Tanh}
	\end{figure}
	
	\ref{fig:hn_vs_Tanh} presents the different curves of Holonorm and Tanh. We can see the similarity and the differences of each of them in a same graphic. 
	
	\begin{table}[ht]
		\centering
		\begin{tabular}{|c|c|c|}
			\hline
			Vector & Holonorm Approx. & Tanh Approx. \\
			\hline
			(1, 2, 3) & (0.2108, 0.4216, 0.6324) & (0.7616, 0.9640, 0.9951) \\
			\hline
			(12, 3, -6) & (0.8136, 0.2034, -0.4068) & (1.0000, 0.9951, -0.9999877) \\
			\hline
			(1, -2, 1) & (0.2898, -0.5796, 0.2898) & (0.7616, -0.9640, 0.7616) \\
			\hline
		\end{tabular}
		\caption{Comparison of Holonorm and Tanh on example vectors}
	\label{approx_value}
\end{table}

In \ref{approx_value}, we see that the Tanh function compresses the values, leading to a loss of information about the original signal. In contrast, Holonorm retains the original signal geometry, preserving the relative magnitudes and directions of the vectors.

\begin{figure}[h!]
	\centering
	\includegraphics[width=0.5\textwidth]{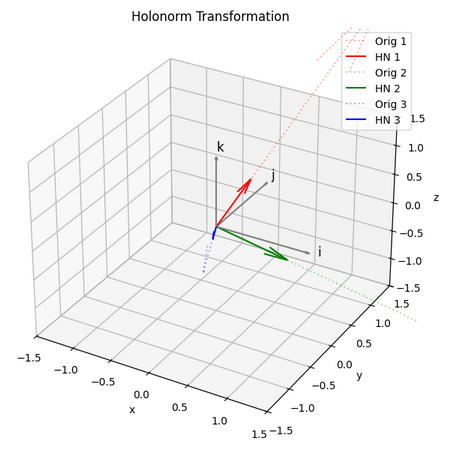}
	\caption{Curves after applying holonorm \\ \\ The above figure \ref{fig:hn orthogonality} represent those 3 vectors after Holonorm and we see that the don't change their direction after the transformation. This mean that Holonorm preserve the meaning of words, the direction and geometry.}
	\label{fig:hn orthogonality}
\end{figure}

\begin{figure}[h!]
	\centering
	\includegraphics[width=0.5\textwidth]{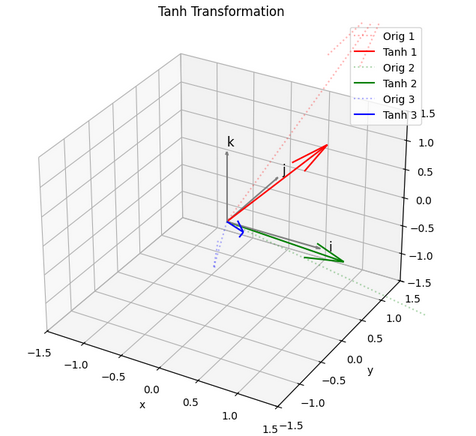}
	\caption{Curves after applying Tanh\\ \\ In \ref{fig:Tanh orthogonality}, the direction of vectors change after the transformation. We clearly see that the angle between vectors change after Tanh. This illustration expresses the fact that Tanh do not preserve the direction and (1, 2, 3), (12, 3, -6), (1, -2, 1) become strongly correlated.}
	\label{fig:Tanh orthogonality}
\end{figure}

They are uncorrelated but become strongly correlated when tanh is applied. In contrast, transformations like holonorm or 
Holonorm/projectivenorm retain the original signal geometry and are thus preferable for applications requiring high-fidelity audio, video, text and time-series representation.\\ 
The reason why we used norm $l_2$ intead of another range of norm is primarily for the default conventional reasons. 

Holonorm is typically defined as:
\[
h_n(x) = \frac{x}{1 + \|x\|}
\]

where \(\|x\|\) is the norm of \(x\). In this context, we will use the L2 norm (Euclidean norm) for the sake of simplicity and consistency,
with the assumption that \(\|x\|\) refers to the L2 norm (Euclidean norm), unless specified otherwise.

This is standard in most machine learning contexts, where:
\[
\|x\|_2 = \sqrt{x_1^2 + x_2^2 + \cdots + x_n^2}
\]

The second reason is the consistency with Tanh. Since Tanh acts element-wise and does not depend on a norm, comparing it to Holonorm using the $L2$ 
norm (standard) makes the comparison balanced and fair.

We can also say that the $L2$ norm is smooth and differentiable everywhere except at zero.

This experiment shows that Tanh should not be used in the context of normalization, as it distorts the original signal and does not preserve the meaning of the data.

\begin{table}[h]
	\centering
	\caption{Comparison of tanh vs. HoloNorm in Signal Processing}
	\begin{tabular}{|p{4cm}|p{5cm}|p{5cm}|}
		\hline
		Property & Tanh & HoloNorm: $hn$ \\
		\hline
		Map & $x \mapsto \tanh(x)$ & $x \mapsto \frac{x}{1+\|x\|} = hn(x)$ \\
		\hline
		Inverse & $\tanh^{-1}(x) = \frac{1}{2} \ln \left( \frac{1+x}{1-x} \right)$ & $y \mapsto \frac{y}{1-\|y\|}$, with $\|y\| < 1$ \\
		\hline
		Derivative & $\frac{d}{dx} \tanh(x) = 1 - \tanh^2(x)$ & $I(1+\|x\|) - \frac{xx^T}{\|x\|(1+\|x\|)^2}$ \\
		\hline
		Primitive & $\ln(\cosh(x)) + C$ & $\|x\| - \ln(1 + \|x\|) + C$ \\
		\hline
		Preserves orthogonality & No & Yes \\
		\hline
		Preserves direction & No & Yes \\
		\hline
		Suitable for audio reconstruction & Poor & Strong \\
		\hline
		Component-wise distortion & Yes & No \\
		\hline
		Induced correlation between signals & Yes & No \\
		\hline
		Bounded output & Yes ($-1, 1$) per component & Yes (open unit ball in $\mathbb{R}^n$) \\
		\hline
	\end{tabular}
	\label{tab:comparison}
\end{table}

\ref{tab:comparison} is the comparison table of the properties of Holonorm and Tanh. 
\subsection{Properties of Normalization}

A Transformer Block consists of a series of operations applied to input data, typically in the context of 
natural language processing or other sequential data tasks. The block is designed to process input tokens in 
parallel, allowing for efficient computation and capturing long-range dependencies.
This is a brief description of a Transformer Block.

Let \( y_{\ell-1,i} \in \mathbb{R}^d \) be the input at layer \( \ell \). For each token \( i \), we compute:

\begin{align} 
	y^{(1)}_{\ell, i} &= h_n(y_{\ell-1, i}) \\
	q_i &= W_q y^{(1)}_{\ell,i}, \quad
	k_j = W_k y^{(1)}_{\ell,j}, \quad
	v_j = W_v y^{(1)}_{\ell,j} \\
	\alpha_{ij} &= \frac{\exp(\langle q_i, k_j \rangle / \sqrt{d})}{\sum_{j'=1}^{i} \exp(\langle q_i, k_{j'} \rangle / \sqrt{d})}, \quad \text{for } j \leq i \\
	\text{Att}_{\ell,i} &= \sum_{j=1}^i \alpha_{ij} v_j \\
	y^{(2)}_{\ell, i} &= y_{\ell-1, i} + \text{Att}_{\ell,i} \\
	y^{(3)}_{\ell, i} &= h_n(y^{(2)}_{\ell, i}) \\
	\text{FF}_{\ell,i} &= W_2 h_n(W_1 y^{(3)}_{\ell,i} + b_1) + b_2 \\
	y_{\ell, i} &= y^{(2)}_{\ell, i} + \text{FF}_{\ell,i} \label{3}
\end{align}

Repeat this process for \( \ell = 1, \dots, L \). The final output is \(\hat{y}_i (\bm{\theta}) = y_{L,i} \).

\section*{Normalization in Transformer Architectures}
Normalization is a fundamental operation in Transformer models, crucial for stabilizing training and ensuring efficient optimization. 
The most widely used technique is \textbf{Layer Normalization (LayerNorm)}, which normalizes each token's feature 
vector \( x \in \mathbb{R}^d \) independently across its feature dimensions. The normalization process involves 
the following steps.

\textbf{Compute the Mean of the input vector:}
\[
\mu = \frac{1}{d} \sum_{i=1}^d x_i
\]

\textbf{Compute the Variance:}
\[
\sigma^2 = \frac{1}{d} \sum_{i=1}^d (x_i - \mu)^2
\]

\textbf{Normalize the Input:}
\[
\hat{x}_i = \frac{x_i - \mu}{\sqrt{\sigma^2 + \epsilon}}, \quad \text{for } i = 1, \dots, d
\]

\textbf{Apply Learnable Scale and Shift Parameters:}
\[
\text{LayerNorm}(x_i) = \gamma_i \cdot \hat{x}_i + \beta_i, \quad \gamma, \beta \in \mathbb{R}^d
\]

Here, \( \epsilon \) is a small constant added for numerical stability (typically \( \epsilon = 10^{-5} \)), and \( \gamma_i, \beta_i \) are learnable parameters that allow the model to rescale and shift the normalized output.

In the Transformer architecture, LayerNorm is applied twice within each encoder or decoder block.

\textbf{After the Self-Attention Sublayer:}
\[
x_i^{(1)} = x_i + \text{SelfAttention}(\text{LayerNorm}(x_i))
\]

\textbf{After the Feedforward Sublayer:}
\[
x_i^{(2)} = x_i^{(1)} + \text{FeedForward}(\text{LayerNorm}(x_i^{(1)}))
\]

Regarding architectural choices, two variants are commonly used.

\textbf{Pre-Normalization (Pre-LN)} — LayerNorm is applied \emph{before} the sublayer:
\[
\text{Block}(x) = x + \text{Sublayer}(\text{LayerNorm}(x))
\]

\textbf{Post-Normalization (Post-LN)} — LayerNorm is applied \emph{after} the residual connection:
\[
\text{Block}(x) = \text{LayerNorm}(x + \text{Sublayer}(x))
\]

The Pre-LN configuration is preferred in modern Transformers due to its improved gradient flow and training 
stability, especially in deep networks. Overall, Layer Normalization ensures that each feature dimension of the 
input maintains a controlled distribution, mitigating issues like exploding or vanishing gradients and enhancing 
convergence speed.

\subsubsection{Holonorm Normalization}
Based on the description of the normalization process described in \cite{math12223506} by Tembine et al., where he clearly 
explains the normalization of tensors with Holonor, we generalize it to vectors in the context of deep learning models.

Holonorm is a normalization operator designed to preserve both the geometry and invertibility of feature representations in high-dimensional Transformer layers. For an input token vector \( X_o \in \mathbb{R}^d \), Holonorm maps it to a normalized form via the transformation:
\[
\hat{\sigma}(X_o) = \frac{X_o}{1 + \sqrt{\langle X_o, X_o \rangle}},
\]
where \( \langle \cdot, \cdot \rangle \) denotes the standard inner product. This operation is applied independently to each token, and the full normalization over a batch \( \{X_1, \dots, X_D\} \) is given by:
\[
\mathcal{O}_{l,n}(X_1, \dots, X_D) = \left( \hat{\sigma}(X_1), \dots, \hat{\sigma}(X_D) \right).
\]
Unlike standard LayerNorm, Holonorm does not subtract the mean or divide by the standard deviation across features; instead, it compresses the input vector into the open unit ball while preserving its direction. Notably, Holonorm is invertible: if \( \hat{\sigma}(X_o) = Z_o \), then the inverse is
\[
X_o = \frac{Z_o}{1 - \sqrt{\langle Z_o, Z_o \rangle}}, \quad \text{for } \langle Z_o, Z_o \rangle < 1.
\]
Furthermore, the transformation is smooth and 1-Lipschitz, with Jacobian
\[
\nabla \hat{\sigma}(X_o) = \frac{I}{1 + \|X_o\|} - \frac{X_o^\dagger X_o}{(1 + \|X_o\|)^2 \|X_o\|},
\]
and the norm of the output is bounded:
\[
\|\hat{\sigma}(X_o)\| \leq \min(1, \|X_o\|).
\]
This normalization can also be interpreted as a weighted differential:
\[
\hat{\sigma}(X_o) = \left(\frac{1}{1 + \|X_o\|}\right) \nabla \left( \frac{1}{2} \|X_o\|^2 \right),
\]
which highlights its scale-aware behavior. Holonorm ensures that the normalized vectors retain their original directional semantics while preventing exploding activations, making it particularly well-suited for deep or multimodal Transformer architectures where vector norm preservation and smooth transformations are essential.

\subsubsection{Tanh normalization}
Tanh-based normalization offers an alternative approach to standard normalization methods (Relu, Gelu) in Transformers by exploiting the bounded, smooth, and saturating nature of the hyperbolic tangent function. For an input token 
vector \( X_o \in \mathbb{R}^d \), we define the normalization as an elementwise mapping using the hyperbolic tangent:
\[
\hat{\tau}(X_o) = \tanh(X_o) = \left( \tanh(x_{o,1}), \dots, \tanh(x_{o,d}) \right).
\]
This function nonlinearly squashes each feature to the open interval \( (-1, 1) \), ensuring bounded activations that mitigate the risk of exploding gradients based optimizers in deep Transformer models. Normalization is applied 
independently to each token within the token vector (componentwise) \label{5}, and the full normalization operator on a sequence \( \{X_1, \dots, X_D\} \) 
is given by:
\[
\mathcal{O}_{t,n}(X_1, \dots, X_D) = \left( \hat{\tau}(X_1), \dots, \hat{\tau}(X_D) \right).
\]

Unlike Holonorm or LayerNorm, Tanh-based normalization does not explicitly scale or normalize the norm of the input vector; instead, it controls the output range through functional saturation. This is particularly useful for stabilizing dynamics in attention and feedforward sublayers, as all components of the output remain within a fixed interval only.

The Tanh function is smooth and differentiable everywhere, with the Jacobian given by:
\[
\nabla \hat{\tau}(X_o) = \operatorname{diag}\left( 1 - \tanh^2(x_{o,1}), \dots, 1 - \tanh^2(x_{o,d}) \right),
\]
which is a diagonal matrix with values in \( (0, 1) \), ensuring that the transformation is contractive. Moreover, the function is odd and monotonic, preserving the sign and order of feature values, which helps retain directional semantics in attention contexts.

The maximum output norm is bounded by:
\[
\| \hat{\tau}(X_o) \| \leq \sqrt{d},
\]
where the bound is tight when all components tend toward \( \pm 1 \). While Tanh-based normalization is not invertible in closed form due to the nonlinearity's saturation, it is highly effective in regularizing Transformer activations in a numerically stable and biologically inspired manner. 

\subsection{Holonorm vs Tanh for normalization}
If there is no \texttt{LayerNorm} in the architecture, then \texttt{tanh} may be ok. Then the problem in self-attention is here:

\[
\texttt{softargmax}(Q\, \texttt{layernorm}(x_1), K\, \texttt{layernorm}(x_2), \dots)
\]

Better to remove completely the layernorm there if one works with \texttt{tanh}.
But if we remove it, it becomes computationally unbounded.
But it works with \textbf{holonorm} and stays bounded.

The attention mechanism in Transformer models relies on the triplet: queries $Q$, keys $K$, and values $V$. The similarity between queries and keys determines how values are weighted and aggregated. This mechanism is mathematically related to the cosine similarity, which measures orientation between vectors.

\subsection{ Scaled Dot-Product Attention}
Given:
\begin{itemize}
	\item $Q \in \mathbb{R}^{n \times d}$: query matrix
	\item $K \in \mathbb{R}^{m \times d}$: key matrix
	\item $V \in \mathbb{R}^{m \times d}$: value matrix
\end{itemize}

The attention output for a single query vector $Q_i$ is computed as:
\begin{equation}
	\text{Attention}(Q_i, K, V) = \sum_{j=1}^{m} \alpha_{ij} V_j
\end{equation}
where
\begin{equation}
	\alpha_{ij} = \frac{\exp\left(\frac{Q_i \cdot K_j}{\sqrt{d}}\right)}{\sum_{k=1}^{m} \exp\left(\frac{Q_i \cdot K_k}{\sqrt{d}}\right)}
\end{equation}

The cosine similarity between two vectors $X$ and $Y$ is defined as:
\begin{equation}
	\cos(\theta) = \frac{X \cdot Y}{|X||Y|}
\end{equation}
This measures the angle between the vectors, not their magnitude.

The dot product in attention can be rewritten as:
\begin{equation}
	Q_i \cdot K_j = |Q_i| \cdot |K_j| \cdot \cos(\theta_{ij})
\end{equation}
If $Q_i$ and $K_j$ are normalized to unit vectors:
\begin{equation}
	Q_i \cdot K_j = \cos(\theta_{ij})
\end{equation}
Thus, scaled dot-product attention becomes a softmax over cosine similarities.

In NLP:
\begin{itemize}
	\item $Q_i$: representation of the current word
	\item $K_j$: representations of context words
\end{itemize}
If $\cos(\theta_{ij}) \approx 1$ and $\theta_{ij} \approx 0$, then high cosine similarity implies semantic closeness.

The attention weights can be viewed as a probability distribution:
\begin{equation}
	\alpha_{ij} = P(j \mid i)
\end{equation}
The attention output becomes:
\begin{equation}
	\text{Attention}(Q_i, K, V) = \mathbb{E}_{j \sim P(\cdot | i)}[V_j]
\end{equation}

When vectors $Q_i, K_j$ are unit vectors, they lie on the unit hypersphere. Attention then aggregates nearby vectors on the sphere weighted by angular proximity (cosine similarity).

\begin{table}[h!]
	\centering
         \caption{View of attention mechanism using cosine similarity.}
	\begin{tabular}{|l|l|l|}
		\hline
		\textbf{Concept} & \textbf{Formula} & \textbf{Meaning} \\
		\hline
		Dot Product & $Q_i \cdot K_j$ & Raw similarity with magnitude \\
            \hline
		Cosine Similarity & $\frac{Q_i \cdot K_j}{|Q_i||K_j|}$ & Angular similarity only \\
            \hline
		Attention Score & $\frac{\exp(Q_i \cdot K_j / \sqrt{d})}{\sum_k \exp(Q_i \cdot K_k / \sqrt{d})}$ & Weighted semantic match \\
		\hline
	\end{tabular}
\end{table}

\textbf{Theorem 1:} Holonorm preserves the sign of the correlation for every problem where similarity is used. 
\paragraph*{Proof:}

For our Holomorphic normalization, we can use the Holonorm function defined as:
\[
\text{similarity} = \frac{X \cdot Y}{\ (1 + |X\|) \, (1 + \|Y\|)}
\]
This is a bounded function that maps the similarity to the open unit ball, preserving the sign of the inert product. So, 
the more 2 vectors are correlated, the more the probability attention choose them. As holonorm preserves the orthogonality, 
the Holonorm function is close to 1 if correlated, otherwise the less they are correlated, the more it is close to 0. 

This function is still bounded between 0 and 1, making it suitable for applications where the output needs to be interpreted as a percentage or a probability.

\textbf{Theorem 2:} Tanh doesn't preserve the sign of the correlation and shouldn't be used where similarity score is required. 

\paragraph*{Proof:}
The Tanh function is defined as:
\[\tanh(x) = \frac{e^x - e^{-x}}{e^x + e^{-x}} \]
The Tanh function compresses the values, leading to a loss of information about the original signal. 

The \textbf{attention mechanism} uses inner products like:
\begin{equation}
	\alpha_{ij} = \frac{\exp(Q_i \cdot K_j / \sqrt{d})}{\sum_k \exp(Q_i \cdot K_k / \sqrt{d})}
\end{equation}

If you apply tanh before the dot product:
\begin{equation}
	\alpha_{ij} = \frac{\exp(\tanh(Q_i) \cdot \tanh(K_j))}{\sum_k \exp(\tanh(Q_i) \cdot \tanh(K_k))}
\end{equation}

Effects:
\begin{enumerate}
	\item \textbf{Norm Suppression}:
	\begin{itemize}
		\item Without tanh: large dot products possible $\rightarrow$ sharper softmax (clear focus).
		\item With tanh: inner products bounded in $(-d, d)$, regardless of magnitude.
		\item This makes attention scores \textbf{less discriminative}.
	\end{itemize}
	
	\subsection*{Example 1: Norm Suppression}
	Without tanh:
	\begin{equation}
		|Q| = 100 \Rightarrow Q \cdot K_1 = 100
	\end{equation}
	With tanh:
	\begin{equation}
		|\tanh(Q)| < \sqrt{d} \Rightarrow \text{Max dot product} \leq d
	\end{equation}
	\textbf{Effect:} Large contrast is lost.

	\item \textbf{Directional Compression}:
	\begin{itemize}
		\item Multiple distinct vectors $x_1, x_2$ with $\|x_1\| \gg \|x_2\|$ will both become \textbf{similar vectors} after tanh.
		\item \textbf{Loss of directional fidelity} in high-dimensional space.
	\end{itemize}
	
	\subsection*{Example 2: Directional Compression}
	Let:
	\begin{equation}
		x_1 = [100, 0], \quad x_2 = [10, 0] \Rightarrow x_1 \ne x_2
	\end{equation}
	But:
	\begin{equation}
		\tanh(x_1) \approx \tanh(x_2) \approx [1, 0]
	\end{equation}
	\textbf{Effect:} Distinct vectors become similar.
	
	\item \textbf{Orthogonality Destruction}:
	\begin{itemize}
		\item Suppose $Q_i \perp K_j \Rightarrow Q_i \cdot K_j = 0$
		\item But $\tanh(Q_i) \cdot \tanh(K_j) \neq 0$ generally because $\tanh$ is \textbf{nonlinear and elementwise}.
		\item So orthogonal semantic vectors become \textbf{non-orthogonal}.
	\end{itemize}
\end{enumerate}

\begin{table}[h!]
	\centering
	\begin{tabular}{|l|l|l|}
		\hline
		\textbf{Effect} & \textbf{Cause} & \textbf{Impact} \\
		\hline
		Norm Suppression & $\tanh(x) \in (-1,1)$ & Dot products bounded \\
            \hline
		Directional Compression & Saturation of tanh & Distinct vectors become similar \\
            \hline
		Orthogonality Destruction & Nonlinear elementwise tanh & Breaks geometric meaning \\
		\hline
	\end{tabular}
	\caption{Orthogonality effects of Tanh} \label{4}
\end{table}

\subsection{Tokenization: Holonorm Vs Tanh }

Tokenization is one central aspect in the training of transformer process. Before any normalization ( LayerNorm, BatchNorm, Tanh or Holonorm) happens inside the Transformer, the raw input vector (text) is first tokenized. In other terms, the vector space is converted into an indexing world where tokens get token IDs. These token IDs are then embedded into vectors via a lookup in an embedding matrix:\\
\[
\text{Embeddings} = E_{\text{token\_id}} \in \mathbb{R}^d
\]
\\
Each token gets a vector of size \( d \), and this forms the input sequence matrix \( X \in \mathbb{R}^{T \times d} \), where \( T \) is the length of sequence.

Once the input embeddings are formed, the Transformer processes them through layers. Normalization is applied at various stages, typically after attention sub-layers and feed-forward layers. 
The normalization function is influenced by the size of the token vectors. Mathematically $dim (E_{\text{token}}) = \sum{Token}$ represents the tokenized vector obtained before the normalization layer.   
To have a more comprehensive understanding of that properties, we should refer to Hilbert space where vectors are infinite.\\

\subsubsection{Tanh}
In table table \ref{4}, we see that \textbf{Tanh}, bounded in the unit ball and Tanh is also componentwise \ref{5}. If the size of the vector dim($E_{\text{token}}$) is too big, we will have $\sum_{i=1}^{1^{6 billion}} 1$ or $\sum_{i=1}^{1^{6 billion}} -1$, then dimension of the Token Embedding vector will still grow and grow as in Hilbert space where the vector are infinite. Tanh is good only in small dimension because $\sqrt{d_1}, \sqrt{d_2}, ..., \sqrt{d_n}$ will be too high for each inputs $\hat{x}_i$
\begin{center}
	\[
	\hat{x}_i = \frac{x_i - \mu}{\sqrt{\sigma^2 + \epsilon}}, \quad \text{for } i = 1, \dots, n
	\]
\end{center}
but normalization bounded only if  $\|\sqrt{d_1}, \sqrt{d_2}, ..., \sqrt{d_n}\|$ is small.

\subsubsection{Holonorm}
Holonorm  is dimension independent. It preserves both the geometry and invertibility of vectors, tensors of high dimension. Whatever the dimension of $E_{\text{token}}$, hn takes each entire token vector because it is not component-wise. 

\begin{center}
	\[
	\text{LayerNorm}(x_i) = \gamma_i \cdot \hat{x}_i + \beta_i, \quad \gamma, \beta \in \mathbb{R}^d
	\]
\end{center}

where $x_i$ is the input vector. The value of the normalization will stay within  the unit ball respecting the principle of the mathematical norm.

\subsection{Holonorm as a percentage}
Holonorm can be interpreted as a percentage, where the output is bounded between 0 and 1. This makes it suitable for applications where the output needs to be interpreted as a percentage or a probability. For example, if we have a value \( x \) that represents a quantity, then \( \text{HN}(x) \) can be interpreted as the percentage of that quantity relative to a maximum value, which is \( 1 + |x| \). This interpretation is particularly useful in scenarios where we want to express the output as a fraction of a whole, such as in forecasting or evaluation metrics. 

These evaluations metric can normalize the error and represent the error in term of percentage because, defined on $\mathbb{R} \xrightarrow{f} [0, 1[$. It can bring back a too large value of metrics such as MAE or RMSE between 0 and 1 without losing any information because the denominator cannot be zero. Thus, it loses no information and allows for the recovery of the original expression. Being between the unit diameter, it can be directly considered as a percentage by multiplying by 100. Consequently, 1-hn(x) is its complement in terms of probability.

\section{Experiments}

\subsubsection{musical dataset}

The MusicCaps dataset comprises 5,521 music samples, each accompanied by an English aspect list and a free text caption authored by musicians. An aspect list, for instance, might read: \"pop, tinny wide hi hats, mellow piano 
melody, high pitched female vocal melody, sustained pulsating synth lead.\" The caption is composed of multiple sentences describing the music, such as: \"A low-sounding male voice is rapping over fast-paced drums playing a 
reggaeton beat along with a bass. Something like a guitar is playing the melody along. This recording is of poor audio quality. In the background, laughter can be noticed.  This song may be playing in a bar.\" The text focuses 
exclusively on describing the music's sound, excluding metadata like the artist's name. The labeled examples are 10-second music clips sourced from the AudioSet dataset, with 2,858 clips from the evaluation set and 2,663 from 
the training set. https://www.kaggle.com/datasets/googleai/musiccaps

\begin{figure}[h!]
	\centering
	\includegraphics[width=0.3\textwidth]{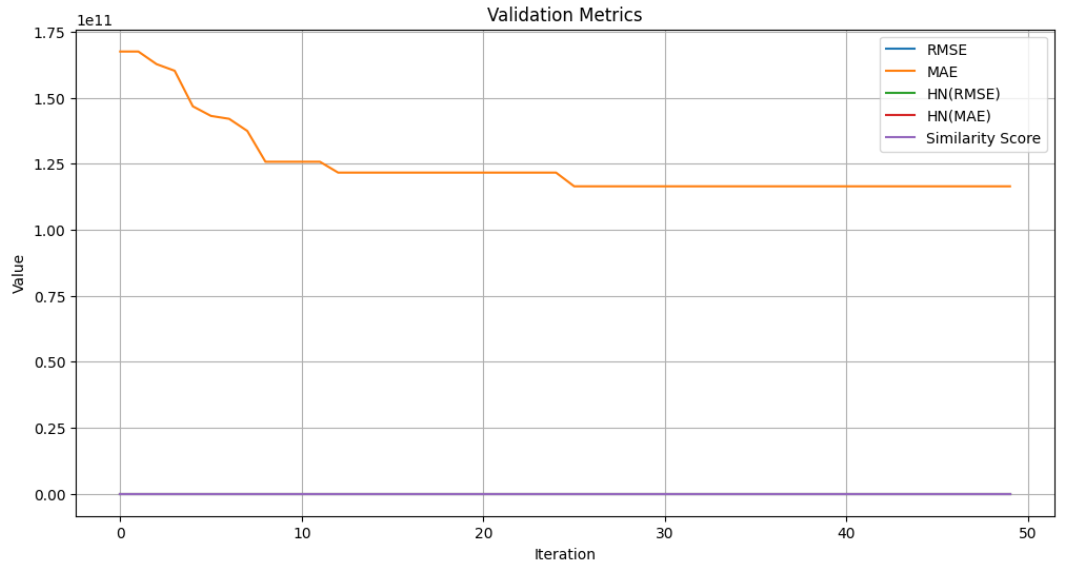}
	\caption{Metrics of music dataset}
	\label{fig: 8}
\end{figure}

In \ref{fig: 8}, the curve of the similarity score stay null along the training. That means vectors preserve their properties.

\newpage
\begin{table}[h!]
	\centering
	\caption{Tanh Performance Metrics for 10 iterations}
	\small
	\begin{tabular}{|p{1cm}|p{2cm}|p{2cm}|p{1.5cm}|p{1.5cm}|p{2cm}|p{2cm}|p{2cm}|}
		\hline
		\textbf{Iteration} & \textbf{Time (s)} & \textbf{Energy Consumption} & \textbf{RMSE} & \textbf{MAE} & \textbf{HN(RMSE)} & \textbf{HN(MAE)} & \textbf{Similarity Score} \\
		\hline
		1 & 756.79 & 7567.87 & 10.8994 & 118.7977 & 0.9160 & 0.9917 & 0.00 \\
		2 & 1513.52 & 15135.22 & 8.7503 & 76.5669 & 0.8974 & 0.9871 & 0.00 \\
		3 & 2267.45 & 22674.51 & 7.2685 & 52.8306 & 0.8616 & 0.9812 & 0.00 \\
		4 & 3015.98 & 30159.82 & 6.9199 & 47.8848 & 0.8737 & 0.9795 & 0.00 \\
		5 & 3768.47 & 37684.70 & 6.9199 & 47.8848 & 0.8737 & 0.9795 & 0.00 \\
		6 & 4517.33 & 45173.32 & 6.6460 & 44.1691 & 0.8692 & 0.9779 & 0.00 \\
		7 & 5287.59 & 52875.92 & 6.2585 & 42.2786 & 0.8635 & 0.9766 & 0.00 \\
		8 & 6054.81 & 60548.06 & 6.3275 & 40.0377 & 0.8665 & 0.9756 & 0.00 \\
		9 & 6830.00 & 68300.00 & 6.1983 & 38.4195 & 0.8631 & 0.9740 & 0.00 \\
		10 & 7609.70 & 76096.99 & 6.1279 & 37.5514 & 0.8597 & 0.9741 & 0.00\\
		\hline
	\end{tabular}
	\label{tab:music hn metrics}
\end{table}

Here, we see that the orthogonality is respected, and the similarity score is 0. This means that the model is not overfitting the data, and the model is able to preserve the information along the training. 
\begin{figure}[h!]
	\centering
	\includegraphics[width=0.3\textwidth]{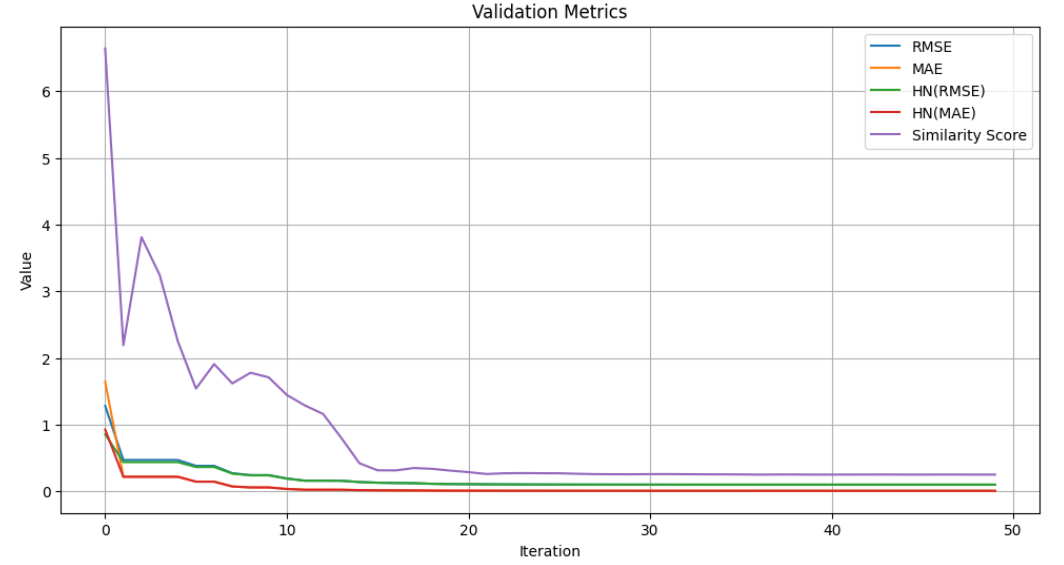}
	\caption{Tanh metrics of music dataset}
	\label{fig: 9}
\end{figure}

Tanh distorts the angles between vectors. In music datasets, the relative positions of vectors are crucial during training, as the resulting sound must be highly harmonized.\\


\begin{table}[h!]
	\centering
	\caption{Tanh Performance Metrics for 10 iterations}
	\small
	\begin{tabular}{|p{1cm}|p{2cm}|p{2cm}|p{1.5cm}|p{1.5cm}|p{2cm}|p{2cm}|p{2cm}|}
		\hline
		\textbf{Iteration} & \textbf{Time (s)} & \textbf{Energy Consumption} & \textbf{RMSE} & \textbf{MAE} & \textbf{HN(RMSE)} & \textbf{HN(MAE)} & \textbf{Similarity Score} \\
		\hline
		1  & 24.55  & 245.47  & 1.2817 & 1.6428 & 0.8569 & 0.9279 & 6.6375 \\
		2  & 47.03  & 470.33  & 0.4720 & 0.2228 & 0.4398 & 0.2192 & 2.1914 \\
		3  & 69.80  & 698.03  & 0.4720 & 0.2228 & 0.4398 & 0.2192 & 3.8103 \\
		4  & 92.13  & 921.32  & 0.4720 & 0.2228 & 0.4398 & 0.2192 & 3.2443 \\
		5  & 114.45 & 1144.54 & 0.4720 & 0.2228 & 0.4398 & 0.2192 & 2.2503 \\
		6  & 136.78 & 1367.80 & 0.3843 & 0.1477 & 0.3664 & 0.1466 & 1.5419 \\
		7  & 158.83 & 1588.28 & 0.3843 & 0.1477 & 0.3664 & 0.1466 & 1.9091 \\
		8  & 181.57 & 1815.73 & 0.2741 & 0.0752 & 0.2675 & 0.0750 & 1.6189 \\
		9  & 203.58 & 2035.83 & 0.2460 & 0.0605 & 0.2411 & 0.0604 & 1.7792 \\
		10 & 226.07 & 2260.72 & 0.2460 & 0.0605 & 0.2411 & 0.0604 & 1.7101 \\
		\hline
	\end{tabular}
	\label{tab:music tanh metrics}
\end{table}

As we see in the figure \ref{fig: 9}, the similarity score challenges during the training, the similarity score is not 0,
which means tanh normalization does not preserve the orthogonality, and the similarity score is not 0. This means that the model is overfitting the data, and the model is not able to preserve the information along the training.\\

We parameterize PSO at 50 particles for PSO optimizer, \(w = 0.5\), \(c1 = 0.5\), \(c2 = 0.5\). 

\newpage
\subsection{Orthogonal vectors dataset for ten terms}
We generated 1000 random vectors in \(\mathbb{R}^3\) and computed the cosine similarity between them.

\begin{figure}[h!]
	\centering
	\includegraphics[width=0.3\textwidth]{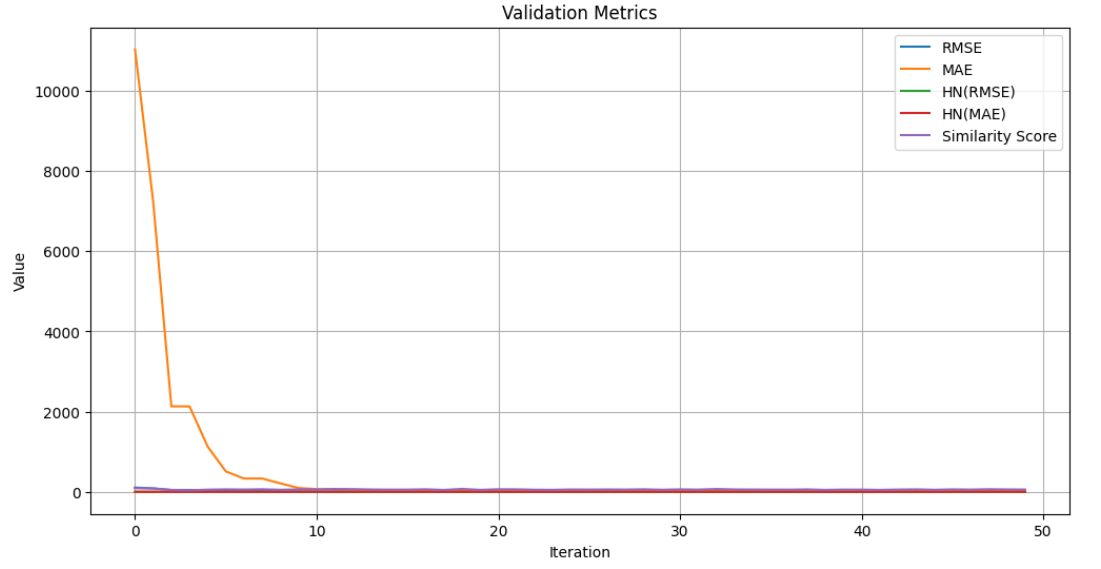}
	\caption{Holonorm metrics of orthogonal vectors dataset}
	\label{fig: 9}
\end{figure}
Here, Holonorm preserves the relationships among vectors, thereby maintaining their orthogonality.\\

\begin{table}[h!]
	\centering
	\caption{Orthogonal HN Performance Metrics for 10 iterations}
	\begin{tabular}{|p{1cm}|p{2cm}|p{2cm}|p{1.5cm}|p{1.5cm}|p{2cm}|p{2cm}|p{2cm}|}
		\hline
		\textbf{Iteration} & \textbf{Time (s)} & \textbf{Energy Consumption} & \textbf{RMSE} & \textbf{MAE} & \textbf{HN(RMSE)} & \textbf{HN(MAE)} & \textbf{Similarity Score} \\
		\hline
		1  & 0.67 & 6.66  & 105.0226 & 11029.7441 & 0.9906 & 0.9999 & 00.00 \\
		2  & 1.20 & 12.05 & 85.0339  & 7230.7698  & 0.9884 & 0.9999 & 00.00 \\
		3  & 1.70 & 17.04 & 46.1627  & 2130.9941  & 0.9788 & 0.9995 & 00.00 \\
		4  & 2.20 & 22.04 & 46.1627  & 2130.9941  & 0.9788 & 0.9995 & 00.00 \\
		5  & 2.75 & 27.49 & 33.5595  & 1126.2416  & 0.9711 & 0.9991 & 00.00 \\
		6  & 3.32 & 33.17 & 22.5328  & 507.7266   & 0.9575 & 0.9980 & 00.00 \\
		7  & 3.67 & 36.74 & 18.2063  & 331.4700   & 0.9479 & 0.9907 & 00.00 \\
		8  & 4.06 & 40.55 & 18.1992  & 331.2109   & 0.9479 & 0.9907 & 00.00 \\
		9  & 4.43 & 44.32 & 14.5199  & 210.8275   & 0.9356 & 0.9953 & 00.00 \\
		10 & 4.80 & 48.00 & 9.7299   & 94.6711    & 0.9068 & 0.9895 & 00.00 \\
		\hline
	\end{tabular}
	\label{tab:ortho_hn_metrics}
\end{table}

The table \ref{tab:ortho_hn_metrics} shows the performance metrics of the orthogonal vectors dataset using Holonorm.
The RMSE and MAE values are significantly lower than those of the music dataset, indicating that the model is able to preserve the information along the training.
\begin{figure}[h!]
	\centering
	\includegraphics[width=0.3\textwidth]{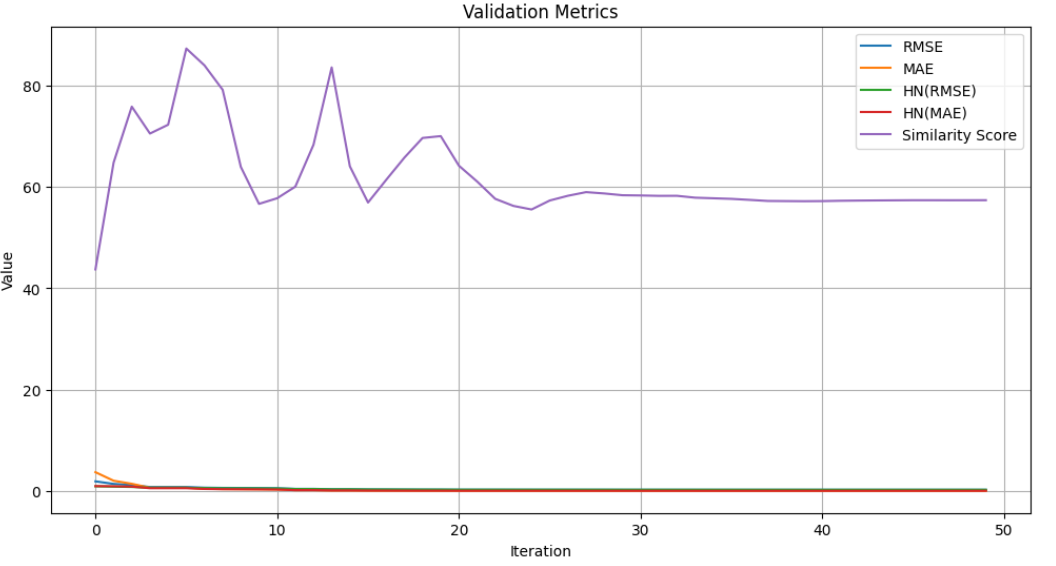}
	\caption{Tanh metrics of orthogonal vector dataset}
	\label{fig: 9}
\end{figure}

Tanh is clearly modify the directions of vectors during the training. \\

\newpage
\begin{table}[h!]
	\centering
	\caption{Tanh orthogonality Metrics for 10 iterations}
	\begin{tabular}{|p{1cm}|p{2cm}|p{2cm}|p{1.5cm}|p{1.5cm}|p{2cm}|p{2cm}|p{2cm}|}
		\hline
		\textbf{Iteration} & \textbf{Time (s)} & \textbf{Energy Consumption} & \textbf{RMSE} & \textbf{MAE} & \textbf{HN(RMSE)} & \textbf{HN(MAE)} & \textbf{Similarity Score} \\
		\hline
		1  & 0.20 & 1.97  & 1.9352 & 3.7451 & 0.9592 & 0.9989 & 43.7253 \\
		2  & 0.32 & 3.25  & 1.4346 & 2.0580 & 0.8926 & 0.9679 & 64.7862 \\
		3  & 0.45 & 4.53  & 1.2071 & 1.4572 & 0.8358 & 0.8971 & 75.8211 \\
		4  & 0.59 & 5.90  & 0.8198 & 0.6721 & 0.6750 & 0.5863 & 70.5129 \\
		5  & 0.72 & 7.19  & 0.8198 & 0.6721 & 0.6750 & 0.5863 & 72.2550 \\
		6  & 0.89 & 8.90  & 0.8198 & 0.6721 & 0.6750 & 0.5863 & 87.2803 \\
		7  & 1.10 & 10.97 & 0.6883 & 0.4737 & 0.5969 & 0.4412 & 83.9534 \\
		8  & 1.29 & 12.87 & 0.6437 & 0.4143 & 0.5674 & 0.3921 & 79.1411 \\
		9  & 1.47 & 14.68 & 0.6299 & 0.3967 & 0.5580 & 0.3771 & 63.9334 \\
		10 & 1.64 & 16.40 & 0.6129 & 0.3756 & 0.5462 & 0.3589 & 56.6497 \\
		11 & 1.83 & 18.30 & 0.5839 & 0.3409 & 0.5255 & 0.3283 & 57.7575 \\
		\hline
	\end{tabular}
	\label{tab:tanh metrics}
\end{table}

The table \ref{tab:tanh metrics} shows the performance metrics of the orthogonal vectors dataset using Tanh normalization. The RMSE and MAE values are significantly higher than those of the Holonorm, indicating that the model is not able to preserve the information along the training.\\

\subsection{Discussion}
The similarity score in this work is defined as the difference between the cosine similarity of a vector before and after normalization. Our analysis shows that Holonorm preserves the direction of the original vector, resulting in no change to the angle between vectors. In contrast, Tanh-based normalization alters the magnitude and direction, leading to distortions in the angular relationship and thus modifying the similarity score.   

\section{Conclusion}
We have shown that the Holonorm function is a better alternative to Tanh for normalization in deep learning models. Today, the GPT-4o model (as used in ChatGPT) supports a context window of up to 128,000 tokens which is something like 500 pages but in the next years, the number of tokens would be billions. We propose Holonorm function that preserves the original signal geometry, maintains orthogonality, and does not distort signals as seen with the cosine similarity score. 
It is also computationally more efficient and numerically stable, making it suitable for applications requiring high-fidelity audio, video, text, and time-series representation.

\section{Proof of Proposition 1} \ref{2}
{\bf Proof:}
	\section*{The Tanh Function}
	
	\subsection*{Definition}
	\[
	\tanh(x) = \frac{\sinh(x)}{\cosh(x)} = \frac{e^x - e^{-x}}{e^x + e^{-x}}
	\]
	
	\subsection*{Inverse of Tanh}
	Let:
	\[
	y = \tanh(x) = \frac{e^x - e^{-x}}{e^x + e^{-x}}
	\]
	Let’s solve for \( x \) in terms of \( y \).
	
	Let:
	\[
	u = e^x, \quad \text{so} \quad e^{-x} = \frac{1}{u}
	\]
	
	Then:
	\[
	y = \frac{u - \frac{1}{u}}{u + \frac{1}{u}} = \frac{u^2 - 1}{u^2 + 1}
	\]
	
	Solving for \( u^2 \):
	\[
	y(u^2 + 1) = u^2 - 1 \implies y u^2 + y = u^2 - 1 \implies (y - 1)u^2 = -(y + 1) \implies u^2 = \frac{y + 1}{1 - y}
	\]
	\[
	\implies e^{2x} = \frac{1 + y}{1 - y} \implies x = \frac{1}{2} \ln\left(\frac{1 + y}{1 - y}\right)
	\]
	
	Thus:
	\[
	\tanh^{-1}(y) = \frac{1}{2} \ln\left(\frac{1 + y}{1 - y}\right)
	\]
	
	\subsection*{Derivative of Tanh}
	\[
	\frac{d}{dx} \tanh(x) = \frac{d}{dx} \left( \frac{e^x - e^{-x}}{e^x + e^{-x}} \right)
	\]
	
	Let’s denote:
	\[
	f(x) = e^x - e^{-x}, \quad g(x) = e^x + e^{-x}
	\]
	
	Then by the quotient rule:
	\[
	\tanh'(x) = \frac{f'g - fg'}{g^2} = \frac{(e^x + e^{-x})(e^x + e^{-x}) - (e^x - e^{-x})(e^x - e^{-x})}{(e^x + e^{-x})^2}
	\]
	
	Calculate numerator:
	\[
	\text{Numerator} = (e^x + e^{-x})^2 - (e^x - e^{-x})^2 = [e^{2x} + 2 + e^{-2x}] - [e^{2x} - 2 + e^{-2x}] = 4
	\]
	
	So:
	\[
	\tanh'(x) = \frac{4}{(e^x + e^{-x})^2} = 1 - \tanh^2(x)
	\]
	
	\section*{Integral of Tanh}
	
	We want to compute:
	\[
	\int \tanh(x) \, dx
	\]
	
	Recall:
	\[
	\tanh(x) = \frac{\sinh(x)}{\cosh(x)}
	\]
	
	Let’s use substitution:
	
	Let \( u = \cosh(x) \implies \frac{du}{dx} = \sinh(x) \).
	
	Then:
	\[
	\int \tanh(x) \, dx = \int \frac{\sinh(x)}{\cosh(x)} \, dx = \int \frac{1}{u} \cdot du = \int \frac{1}{u} \, du = \ln|\cosh(x)| + C
	\]
	
	Result:
	\[
	\int \tanh(x) \, dx = \ln|\cosh(x)| + C
	\]
	
	\section*{Holonorm Function}
	
	\subsection*{Definition}
	\[
	h_n(x) = \frac{x}{1 + |x|}
	\]
	
	\subsection*{Inverse of Holonorm}
	Let:
	\[
	y = \frac{x}{1 + |x|}
	\]
	
	We separate into two cases:
	
	\begin{itemize}
		\item \textbf{Case 1:} \( x \geq 0 \implies |x| = x \)
		\[
		y = \frac{x}{1 + x} \implies y(1 + x) = x \implies y + yx = x \implies y = x(1 - y) \implies x = \frac{y}{1 - y} \quad (\text{provided } y < 1)
		\]
		
		\item \textbf{Case 2:} \( x < 0 \implies |x| = -x \)
		\[
		y = \frac{x}{1 - x} \implies y(1 - x) = x \implies y - yx = x \implies y = x(1 + y) \implies x = \frac{y}{1 + y} \quad (\text{provided } y > -1)
		\]
	\end{itemize}
	
	So the inverse is piecewise:
	\[
	h_n^{-1}(y) =
	\begin{cases}
		\frac{y}{1 - y} & \text{if } y \geq 0 \\
		\frac{y}{1 + y} & \text{if } y < 0
	\end{cases}
	\]
	
	\subsection*{Derivative of Holonorm}
	We again consider piecewise definition of \( h_n(x) \):
	
	\begin{itemize}
		\item \textbf{Case 1:} \( x > 0 \)
		\[
		h_n(x) = \frac{x}{1 + x} \implies h_n'(x) = \frac{(1 + x)(1) - x(1)}{(1 + x)^2} = \frac{1}{(1 + x)^2}
		\]
		
		\item \textbf{Case 2:} \( x < 0 \)
		\[
		h_n(x) = \frac{x}{1 - x} \implies h_n'(x) = \frac{(1 - x)(1) - x(-1)}{(1 - x)^2} = \frac{1 + x}{(1 - x)^2} = \frac{1}{(1 - x)^2}
		\]
		
		At \( x = 0 \): Derivative is continuous.
	\end{itemize}
	
	So:
	\[
	h_n'(x) =
	\begin{cases}
		\frac{1}{(1 + x)^2} & x > 0 \\
		\frac{1}{(1 - x)^2} & x < 0 \\
		1 & x = 0
	\end{cases}
	\]
	
	\section*{Integral of Holonorm (Scalar)}
	We consider both sides (piecewise).
	
	\subsection*{Case 1: \( x \geq 0 \implies h_n(x) = \frac{x}{1 + x} \)}
	Use substitution:
	
	Let \( u = 1 + x \implies du = dx, \quad x = u - 1 \).
	
	Then:
	\[
	\int \frac{x}{1 + x} \, dx = \int \frac{u - 1}{u} \, du = \int \left(1 - \frac{1}{u}\right) du = u - \ln|u| + C \implies (1 + x) - \ln(1 + x) + C
	\]
	
	Result for \( x \geq 0 \):
	\[
	\int \frac{x}{1 + x} \, dx = (1 + x) - \ln(1 + x) + C
	\]
	
	\subsection*{Case 2: \( x < 0 \implies h_n(x) = \frac{x}{1 - x} \)}
	
	Let \( u = 1 - x \implies du = -dx, \quad x = 1 - u \).
	
	Then:
	\[
	\int \frac{x}{1 - x} \, dx = \int \frac{1 - u}{u} \cdot (-du) = \int \left(\frac{u - 1}{u}\right) du = -\int \left(1 - \frac{1}{u}\right) du = -(u - \ln u) + C \implies 
   \]
    $-(1 - x - \ln(1 - x)) + C = x + \ln(1 - x) + C$
	
	Result for \( x < 0 \):
	\[
	\int \frac{x}{1 - x} \, dx = x + \ln(1 - x) + C
	\]
	
	\subsection*{Final Expression for the Integral of HN}
	\[
	\int h_n(x) \, dx =
	\begin{cases}
		(1 + x) - \ln(1 + x) + C & \text{if } x \geq 0 \\
		x + \ln(1 - x) + C & \text{if } x < 0
	\end{cases}
	\]
	
	\section*{Comparison Table}
	
	\begin{tabular}{|l|l|l|l|}
		\hline
		Property & Tanh & Holonorm (HN) & Verdict \\
		\hline
		Expression & \(\frac{e^x - e^{-x}}{e^x + e^{-x}}\) & \(\frac{x}{1 + |x|}\) & \\
		\hline
		Inverse & \(\frac{1}{2} \ln\left(\frac{1 + y}{1 - y}\right)\) & \(\frac{y}{1 \pm y}\) depending on sign & Holonorm faster \\
		\hline
		Derivative & \(1 - \tanh^2(x)\) & \(\frac{1}{(1 \pm x)^2}\) & Holonorm cheaper \\
		\hline
		Numerical Stability & Poor near \( x \gg 1 \) & & Holonorm wins \\
		\hline
		Computational Cost & Exponentials, logarithms & No exp/log, only division and square & Holonorm wins \\
		\hline
	\end{tabular}

\end{document}